\documentclass[sn-mathphys]{sn-jnl}% Math and Physical Sciences Reference Style
\normalbaroutside

\usepackage{color}
\usepackage{mathtools}
\usepackage{tikz}
\usepackage[ruled,vlined,algo2e]{algorithm2e}

\jyear{2021}%
\def\WORKDIR{.}

\newcommand{\layerAxeSize}{\mathcal{L}}
\newcommand{\scaleAxeSize}{\mathcal{S}}
\newcommand{\channelAxeSize}{\mathcal{C}}
\newcommand{\resWidth}{W}
\newcommand{\resHeight}{H}

\newcommand{\loss}{\mathcal{L}}
\newcommand{\prunedElems}{\mathcal{P}_{E}}

\newcommand{\prunedLinks}{\mathcal{P}_{L}}
\newcommand{\cond}{\mathcal{C}}
\newcommand{\elems}{\mathcal{E}}
\newcommand{\crit}{\chi}
\newcommand{\mask}{\mathcal{M}}
\newcommand{\dataset}{D}
\newcommand{\entity}{x}
\newcommand{\clabel}{y}
\newcommand{\plabel}{\Tilde{\clabel}}
\newcommand{\elabel}{\hat{\clabel}}
\newcommand{\process}{E}

\newcommand{\normE}[1]{\left\lVert #1 \right\rVert}

\theoremstyle{thmstyleone}%
\theoremstyle{thmstyletwo}%

\theoremstyle{thmstylethree}%

\begin{document}

\title[Convolutional Network Fabric Pruning With Label Noise]{Convolutional Network Fabric Pruning With Label Noise}

% \author{Ilias Benjelloun}\affiliation{Université de Lorraine, 
% CNRS, LORIA, F-54000 Nancy, France
% }

% \author{Efoevi Angelo Koudou}
% \affil[1]{Université de Lorraine, 
% CNRS, IECL, F-54000 Nancy, France
% }

% \author{Bart Lamiroy}
% \affiliation{Université de Reims Champagne Ardenne, CReSTIC EA 3804, F-51100 Reims, France}

\author*[1]{\fnm{Ilias} \sur{Benjelloun}}\email{ilias.benjelloun@mines-nancy.org}

\author*[2,1]{\fnm{Bart} \sur{Lamiroy}}\email{bart.lamiroy@univ-reims.fr}
\equalcont{These authors contributed equally to this work.}

\author[3]{\fnm{Efoevi Angelo} \sur{Koudou}}
\equalcont{These authors contributed equally to this work.}

\affil*[1]{\orgname{Université de Lorraine, 
CNRS, LORIA}, \orgaddress{\postcode{F-54000}, \city{Nancy}, \country{France}}}

\affil*[2]{\orgname{Université de Reims Champagne Ardenne, CReSTIC EA 3804}, \orgaddress{\postcode{F-51100}, \city{Reims}, \country{France}}}

\affil[3]{\orgname{Université de Lorraine, 
CNRS, IECL}, \orgaddress{\postcode{F-54000}, \city{Nancy}, \country{France}}}

% \affil[2]{\orgdiv{Department}, \orgname{Organization}, \orgaddress{\street{Street}, \city{City}, \postcode{10587}, \state{State}, \country{Country}}}

% \affil[3]{\orgdiv{Department}, \orgname{Organization}, \orgaddress{\street{Street}, \city{City}, \postcode{610101}, \state{State}, \country{Country}}}


\abstract{This paper presents an iterative pruning strategy for Convolutional Network Fabrics (CNF) in presence of noisy training and testing data. 
	
	With the continuous increase in size of neural network models, various authors have developed pruning approaches to build more compact network structures requiring less resources, while preserving performance. As we show in this paper, because of their intrinsic structure and function, Convolutional Network Fabrics are ideal candidates for pruning. 
	
	We present a series of pruning strategies that can significantly reduce both the final network size and required training time by pruning either entire convolutional filters or individual weights, so that the grid remains visually understandable but that overall execution quality stays within controllable boundaries. Our approach can be iteratively applied during training so that the network complexity decreases rapidly, saving computational time. The paper addresses both data-dependent and data-independent strategies, and also experimentally establishes the most efficient approaches when training or testing data contain annotation errors.}




\maketitle

\section{Introduction}\label{sec:intro}

Deep learning has been successful in solving very hard problems, but in the same time, the complexity of underlying neural networks has not ceased to increase. The hardware architectures required for achieving state-of-the-art performance are not necessarily available to all, or in constrained real-world situations like on hand-held devices. This is why research on pruning has gained momentum. Network pruning consists of determining parts of neural networks that do not significantly contribute to their final performance, and removing them. It has been first investigated in~\cite{lecun,hanson,hassibi}. Pruning was mostly applied on pre-trained networks, followed by fine-tuning. More recently, other approaches have emerged, like iteratively alternating between training and pruning~\cite{carreira}, or even pruning networks immediately after they have been initialized, so that one only needs to train a sparse architecture~\cite{frankle,zhou,ramanujan}.

Furthermore, since designing the most appropriate neural net architecture for a given problem requires time, knowledge and experience for selecting and setting many hyperparameters, Convolutional Neural Fabrics (CNF) have been introduced~\cite{vebreek} to facilitate the design of convolutional networks. CNFs embed an exponentially large number of chain-structured convolutional networks in one grid-like architecture, where a linear path from the input to the output of the grid corresponds to one CNN over many possible architecture choices. A CNF is defined by only two hyperparameters, and can be trained with the traditional backpropagation technique. Therefore, it facilitates the model selection process, since training the CNF is roughly similar to training multiple CNNs, while relying on a smaller number of hyperparameters. Although the CNF architecture is still huge in terms of the number of parameters, it still requires less resources compared to training each embedded CNN alone.

To the best of our knowledge, pruning has not seriously been investigated for CNFs. \cite{vebreek} shows how a pre-trained CNF can be pruned and fine-tuned with minor loss of accuracy, but this implies training a huge architecture during many epochs, which is time-consuming and requires big computational power. Our first contribution is to successfully apply an iterative pruning algorithm to a CNF, rapidly reducing the number of parameters to train. Furthermore, pruning is rather done on entire convolutional filters, so that the possibility to visualize the most influential paths in the network is not lost. This is an important asset of CNFs since it allows to select the most appropriate chained convolution architectures for a given problem.

A second contribution of this paper is to investigate the influence of data quality, and more specifically label or annotation noise on the pruning. Data is not always available with good quality annotations, and even with a good quality annotation process, inter-annotator agreement can vary, since the annotation problem is inherently ill-defined and partially subjective~\cite{lamiroy:hal-01057362}. We study data-dependant {\itshape vs.} data-independent pruning criteria and how they influence the final quality of resulting networks.

Thus this work offers a way to build good convolutional models even for people:
\begin{itemize}
    \item with a low level of knwoledge, as the CNF network structure needs only two hyperparameters to be defined, which is far simpler than traditional CNN;
    \item with low-quality resources: that is why we consider the data can be of low quality, i.e. noisy, and that the network must be pruned to ease its storage. Furthermore, all our experiments have been done using only one GPU, to ensure the feasability of the method with a low-quality architecture, and for that reason, we will not compare to other works done with multiple GPU.
\end{itemize}

The rest of the paper is organised as follows: the next section presents an overview of related work. Section~\ref{sec:B} describes the pruning process and the choices made at various levels of the experimental design. The obtained results are presented and discussed in Section~\ref{sec:R}, and the last section concludes and presents some perspectives for improvement.

\section{Background}\label{sec:RW}

Deep Neural Networks have largely grown in size to reach current state-of-the-art performance. With the improvement of hardware architectures, and large datasets available for many common tasks, it has become easier to build a network with good performance by using an over-parameterized model. This results in a network with high redundancy, trained and stored with unnecessary computation energy and memory consumption~\cite{denil}.

\subsection{Network Pruning}\label{sec:NP}

To reduce the size of a network, one solution is to apply network pruning~\cite{lecun,hanson,hassibi}. Network pruning corresponds to determining the weights of a network that contribute less to the output performance using various criteria, and remove them from the network. It is applied either on already trained networks~\cite{hassibi,lecun}, and followed by fine-tuning, or in an iterative fashion~\cite{carreira}, pruning some weights after enough training epochs and repeating the process, or else directly at the initialization of the network, which is thus trained for the first time after part of its weights have been pruned~\cite{hayou2020pruning}.

Recent advances have extended previous pruning strategies: \cite{han} applies a pruning algorithm without loss in accuracy of the initial network; \cite{frankle} shows the existence of a sub-network that, when trained in isolation, can perform as well as the whole trained network. Both use weight magnitude values for removing elements. On the other hand, \cite{zhou,ramanujan} focus on finding sub-networks with good performance in a randomly initialized untrained network.

There are mainly three criteria to decide which weights can be pruned. The actual number of weights to prune is usually a user-defined parameter.
%The criterion is applied to get a ranking of the weights and $n$ weights are pruned, $n$ being determined by a user-defined threshold on the values obtained with the criterion, or by the desired sparsity of the pruned network. The criteria are:
\begin{itemize}
	\item Magnitude-based or Zero-shot pruning: each weight is ranked based on its absolute value $|w|$. This criterion is data-independent, \textit{i.e.} it is possible to compute without using any data.
	\item Sensitivity-based or Single-shot pruning~\cite{lee2018snip}: each weight is ranked based on $|w\frac{\partial \loss}{\partial w}|$, where $\loss$ is the loss of the network. This criterion is data-dependent as we need some data samples to compute that loss.
	\item Hessian-based pruning~\cite{wang2020picking}: the weights are selected using the hessian of the loss function. This criterion is also data-dependent.
\end{itemize}

\subsection{Convolutional Neural Fabric}

Designing an architecture by hand adapted to solving a given problem can be difficult given the high number of hyperparameters, and can lead to a time-consuming "test and retry" process. In~\cite{vebreek}, the authors focus on CNNs and propose a multidimensional model for image-related tasks. The model embeds an exponential number of chain-structured CNNs, which can be trained together at once by locally sharing weights. This simplifies the design and selection process of the best CNN architecture, but implies an increase in the size of the model to be trained. Thus, by applying network pruning techniques, the training time and memory storage of the model can be significantly reduced, allowing to be stored and run with modest hardware architectures.

\begin{figure}[t]
	\centering
	\includegraphics[scale=0.65]{\WORKDIR/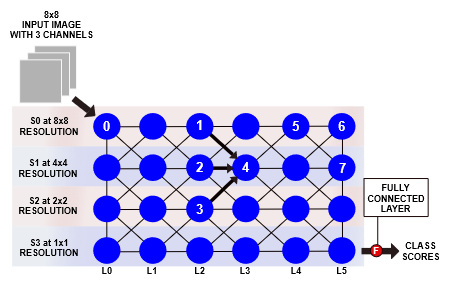}
	\caption{A CNF architecture with 6 layers and 4 scales. Data passes through a first convolution to the input node in the top-left, which propagates it to the other scales by downsampling convolutions. The information flows through the grid along the links, toward the output node (bottom-right) and is passed to a fully connected layer.}
	\label{cnf}
\end{figure}

\begin{figure}[t]
	\centering
	\noindent\begin{tikzpicture}[every node/.style={scale=0.9}]
	\def\hspacing{5.5}
	\def\vspacing{-4.5}
	\def\imscale{0.60}
	\node[label=above:{A}] (cifar10) at (0,0) {\includegraphics[scale=\imscale]{\WORKDIR/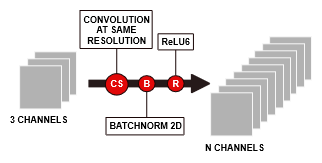}};
	
	\node[label=above:{B}] (cifar100) at (\hspacing,0) {\includegraphics[scale=\imscale]{\WORKDIR/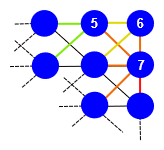}};
	
	\node[label=below:{C}] (svhn) at (0.4*\hspacing,\vspacing) {\includegraphics[scale=\imscale]{\WORKDIR/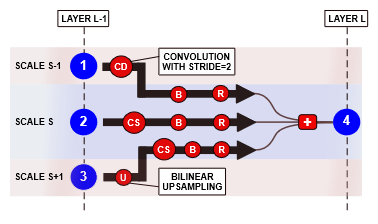}};
	
	\end{tikzpicture}
	\caption{Fig. A and C show how the activation tensor of node 0 is obtained from the input image, and how the activation tensor of node 4 is obtained from nodes 1, 2 and 3. Going from scale $S-1$ to $S$ is done by setting the convolution stride to 2, while the opposite is done through bi-linear upsampling. Fig.~B shows how parts of a CNF become obsolete (green, yellow and orange links) when particular links are pruned (red).}
	\label{cnf_details}
\end{figure}

A CNN architecture is determined by many hyper-parameters: number of convolutional layers ($ConvL$), their number of channels, filter size and stride for each $ConvL$, number of pooling layers ($PoolL$), their operator type and region size as well as the ordering between $ConvL$s and $PoolL$s, channel connectivity pattern between layers, and activation function types per layer. Exhaustive exploration of all possible architectures is not feasible in practice, and finding a good CNN (\textit{e.g.} using a local-search strategy) remains a tough, time-consuming problem when one has to train and test individual architectures.

As described in~\cite{vebreek}, the set of architectures corresponding to all hyperparameter choices can be embedded in a single three-dimensional grid-like structure, called a Convolutional Neural Fabric, and trained in one run. A CNF processes the information along three axes: a layer axis of size $\layerAxeSize$, a scale or resolution axis of size $\scaleAxeSize$, and a channel axis of size $\channelAxeSize$. Since we do not change the number of channels inside a CNF, and we use a standard connectivity pattern between channels, the behaviour of our CNFs does not change across the channel axis. Therefore we will consider only a 2D version of a CNF, with a horizontal layer axis and a vertical scale axis (Fig.~\ref{cnf}). The number of scales is set in function of the size of the input image such that the resolution is divided by 2 when going one scale down, and is equal to $1\times1$ at the smallest scale $\scaleAxeSize-1$.

The input image is first transformed to the right number of channels by a starting convolution operation (Fig.~\ref{cnf_details}.A). The result becomes the activation tensor at the input node of the grid, at position $(layer=0, scale=0)$ in the top-left corner. It is processed until reaching the output node at  $(\layerAxeSize-1, \scaleAxeSize-1)$. In the 2D grid, each node represents an activation tensor of size $(\channelAxeSize, \resWidth, \resHeight)$, where $(\resWidth, \resHeight)$ is the local resolution of the processed data. The  resolution at scale $0$ is the initial resolution of the input image, and decreases along the scale axis. Nodes are connected by links through a local connectivity pattern: node at position $(l, s)$ is linked to the three closest nodes in the previous layer, \textit{i.e.} at positions $(l-1, s \pm 1)$, and the three closest nodes in the next layer, at positions $(l+1, s \pm 1)$ wherever possible. Nodes at layers $0$ and $\layerAxeSize - 1$ must also spread the information along the scale axes, so \textit{e.g.} node $(l,s)$ is connected to node $(l, s+1)$ for $l \in \{0, \layerAxeSize\}$ when possible. Each link applies specific operations to the activation tensor of the node at its start, and passes the resulting tensor to the node at its end. Each node aggregates the tensors it receives to form its activation tensor (Fig.~\ref{cnf_details}.C). For example, the activation tensor $T$ of a node $(l,s)$ in the middle of the grid is computed by $T_{l,s} = \sum_{i \in \{-1,0,1\}} T_{l-1, s+i}$.

%\begin{equation}
%T_{l,s} = \sum_{i \in \{-1,0,1\}} T_{l-1, s+i}
%\end{equation}{}

The typical operations applied by a link are:
\begin{itemize}
	\item a convolution, with a $3\times3$ kernel size, followed by an upsampling or a downsampling according to the direction of the link (up or down), or nothing if the link processes the information at the same scale;
%	\item the upsampling  is a bilinear interpolation and scales activation tensors from a start resolution to an higher end resolution;
%	\item the downsampling allows for the opposite, and is implemented by choosing a stride such that each channel in the output tensor has a size divided by 2;
	\item a batch normalization;
	\item the activation function (in our case ReLU6).
\end{itemize}

In their paper \cite{vebreek}, the authors argue that the only determinant hyperparameters of a CNF are the number of channels and layers, and they show that those two hyperparameters become less critical if their values are large enough.
%, by providing results of CNFs of different sizes trained on three different datasets: CIFAR10, CIFAR100 and ImageNet.

\section{Pruning Algorithm}

In what follows, we denote $\crit(e)$ the pruning criterion applied to a potentially prunable element $e$ of the network. It returns a score expressing how much $e$ contributes to the network performance. We are not interested in pruning at initialization, and focus on zero-shot and single-shot pruning. We divide our process in two steps. 

First, we prune entire links from the CNF, removing them from the flow of information in the fabric. For example, in Fig.~\ref{cnf}, if the link between nodes $3$ and $4$ is pruned, node $4$ only receives two tensors to sum instead of three. This way we first determine an optimal substructure in the fabric identifying the most efficient processing paths for the problem at hand. Second, we apply pruning to the weights in the convolution filters of each remaining link in the fabric. Furthermore, and for obvious reasons, the first link transforming the input image into the right number of channels (Fig.~\ref{cnf_details}), as well as the fully connected layer at the end of the network are never pruned.

\subsection{Link Pruning}

During the first step of pruning, removing one link can make other computations in the fabric unnecessary. For example, in Fig.~\ref{cnf_details}.B, the removal of the red link prevents node $7$ to transmit its activation to the following nodes since its sole output link disappears. This means all computations  for the activation tensor of node $7$ become unnecessary. Therefore, removing the red link implies removing the input links of node $7$ (in orange) as well. Since the output links of node $6$ are part of these orange links, the same situation is repeated: the yellow links must also be removed, followed by the green ones, as node $5$ becomes obsolete, too.

We also need to avoid that the removal of a link entirely cuts the information flow between the input and the output of the fabric. To avoid this situation, we ensure that the set of potentially pruned links $\prunedLinks$ satisfies the condition that if all links in $\prunedLinks$ are removed from the fabric, there still exists a path between the input and output nodes. We achieve this by first ranking the links based on their respective criterion value, then adding links one by one to the set, constantly checking above mentioned condition. Links violating the condition are skipped. We stop when enough links have been found, or when all links have been seen.
%Another possible situation during link pruning is the removal of a link that globally cut the information flow between the input and the output of the fabric. It happens for instance if we want to remove the three output links of node $0$ (in Fig.~\ref{cnf}). To avoid that situation, we must ensure that the set $\prunedLinks$ of links to prune that we find will satisfy the following condition $\cond(\prunedLinks)$: if all links in $\prunedLinks$ are removed, there still exists a path between the input and output nodes of the fabric. To find a set $\prunedLinks$ containing $N$ unimportant links, satisfying $\cond(\prunedLinks)$, without having to greedily check all possible combinations of $N$ links in the fabric, we first rank the links based on their respective criterion value, then we add links one by one in the set, constantly checking condition $\cond(\prunedLinks)$. When adding a link makes the condition false, we simply skip it. We stop when enough links have been found, or if all links have been seen.

The question that remains is how a pruning criterion can be applied to a link. Indeed, they are initially defined to apply on the weights (cf. section \ref{sec:NP}), but links are complex objects. However, as each link contains a weight matrix, the pruning criteria can be adapted in a straightforward way. For a link $l$ and the weight matrix of its convolution filter $W^{l}_{conv}$, the criterion $\crit(l)$ is the euclidean norm of the matrix obtained by applying $\crit$ element by element to $W^{l}_{conv}$: $\crit(l) = \normE{\crit(W^{l}_{conv})}$.
%
%\begin{equation*}
%\crit(l) = \normE{\crit(W^{l}_{conv})}
%\end{equation*}

\subsection{Weight Pruning}

The second step is similar to the first, but pruning now applies to the weights in the convolution filters of the remaining links. In practice, to remove a weight $w$ from the weight matrix $W$, we apply a binary mask $\mask_{W}$ on $W$ with a zero at the same position as $w$, as to cancel any operation involving that weight. This is a convenience solution and will have an impact on the subsequent measurements we can do later on (cf. section \ref{sec:PS}).

Like for Link Pruning, we apply a condition to the weights to determine if we can safely remove them: a weight $w$ cannot be removed if its mask $\mask_{W}$ would result in containing only zeros. We do not want to zero out an entire convolution filter, since the equivalent is already done in the first step.

\begin{algorithm}[]
\SetAlgoLined
\SetKwInOut{Input}{input}
\SetKwInOut{Output}{output}
\Input{$\elems$: a set of fabric elements to potentially prune; either the links of the fabric, or the weights of the convolution filters;\\
$n$: the number of elements to prune;\\
$\crit$: the criterion used to rank the elements by order of importance;\\
$\cond$: condition applied on an element to ensure it can be pruned safely.}
\Output{A set $\prunedElems$ of elements to prune}
An empty list $L$ to store the ranked elements;\\
\For{$e$ in $\elems$}{
Compute the score $\crit(e)$ of $e$;\\
Insert $e$ in $L$ so that $L$ is ranked according to $\crit(e)$;}

Set $\prunedElems$ to empty;\\
\For{$e$ in $L$}{
\If{$\cond(e)$}{
Add $e$ to $\prunedElems$;}
}
 \caption{Pruning algorithm}
\end{algorithm}

\section{Experimental Setup}\label{sec:B}

\subsection{Data}

We consider 4 different datasets for our experiments:
\begin{itemize}
    \item The CIFAR10 dataset~\cite{krizhevsky2009learning}: 60,000 $32\times32$ RGB images, labeled in 10 classes, each containing 6.000 images. There are 5,000 training images and 1,000 test images per class.
    \item The CIFAR100 dataset~\cite{krizhevsky2009learning}: similar to the previous one, but having 100 classes, with each 500 training images and 100 test images.
    \item The SVHN dataset~\cite{netzer2011reading}: 600,000 $32\times32$ RGB images of digits. We only use the 73,257 images of the training set and the 26,032 of the testing, excluding the 531,131 extra training images. 
    \item The VOC2012 classification dataset of the PASCAL VOC challenge~\cite{pascal-voc-2012}: 11,530 RGB images of 500 pixels in width and various heights, with a total of 27,450 ROI annotated objects and 6,929 segmentations. There is a total of 20 classes. Each image can contain multiple objects, and the annotation includes information about each object (class, bounding box, {\itshape etc.}). For that reason, the classification challenge is a multilabel problem (the goal is to find all the objects present in each image). As we only focus on one-label classification, we first apply a preprocessing step to collect only the images that are suited for that task. This way, we collect 9,340 samples out of 11,530.

Preprocessing for each item $d$ in the dataset is as follows:
\begin{itemize}
    \item if $d$ contains only one object or many objects of the same type, keep $d$ with the corresponding object label;
    \item otherwise compute the area occupied by each different object using their bounding boxes found in the annotation of $d$; if the object with the largest surface is twice as big as the object with the second largest surface, keep $d$ with the label of the bigger one, else discard $d$.
\end{itemize}
\end{itemize}

\subsection{Protocol}

In order to show how our CNF pruning algorithm behaves, we compare the performance of an unpruned baseline CNF with differently pruned CNFs trained on the same data. This gives us 4 baseline CNFs (one for each dataset) and 18 differently pruned CNFs for each baseline. Experiments are run on both the standard datasets ({\itshape cf.} Section~\ref{sec:R}) and on the same datasets in which we artificially introduce labelling errors ({\itshape cf.} Section~\ref{sec:noise}), so that we can assess how the mislabeling may influence the pruning process. We implemented our experiments in PyTorch, and we trained the models on the Grid5000 platform~\cite{grid5000}, using one node (physical machine) with one GPU per training. %Results are reported in Fig.~\ref{clean-1} and~\ref{clean-2}.

\subsubsection{Training Data}
  systematically consists in only 90\% of the provided training sets (for those datasets providing training and test sets). The remaining 10\% is kept for validation. The test set is used only for gradient computation when one-shot pruning is applied ({\itshape cf.}~\ref{exp-pruning}). We always make sure to split the data such that CNFs are consistently trained on data sets of the same size, independently of the pruning criterion. As the PASCAL VOC challenge does not provide any test set, we apply the following particular split for its data: 70\% of the data is used for training, 20\% for testing, and the remaining 10\% for validation. When splitting the datasets, we keep the same proportion of each class in each split. The data augmentation scheme we apply is generally composed of normalizing, resizing, randomly cropping and flipping horizontally the images (Fig.~\ref{augment} in appendix). 

\subsubsection{CNF Architecture} consistently has 8 layers and 64 channels for all but one dataset: the SVHN dataset. Our experiments have shown that 64 channels is too much for that particular problem. We reduced the number of channels to 32 when training on SVHN. We use a cross entropy loss for training, and we optimize the networks with stochastic gradient descent during 200 epochs, using an initial learning rate of 0.1, and dividing it by 10 once after epoch 80, and a second time after epoch 120.

\subsubsection{Pruning\label{exp-pruning}}
\begin{enumerate}
    \item We set three desired sparsity values of the final network: 0.05, 0.03, 0.01; a sparsity of 0.05 means that we remove 95\% of the elements in the CNF (95\% of links and 95\% of weights); we also always ensure that the number of remaining links is greater than the number of links contained in the longest linear path in the CNF, no matter the sparsity specified: this is for letting a chance to at least each linear path to be kept during link pruning;
    \item We apply three different strategies;
    \begin{itemize}
        \item early pruning: after 5 epochs of training, we apply link pruning, followed by weight pruning to reach the desired sparsity;
        \item late pruning: after 75 epochs of training, we apply link pruning, followed by weight pruning to reach the desired sparsity;
        \item iterative pruning: after 5 epochs of training, we apply link pruning followed by weight pruning, and this every 10 epochs until epoch 75; the number of links and weights pruned each time is calculated so that at the end, the right number has been removed according to the sparsity value specified;
    \end{itemize}{}
    \item For each strategy, we both test magnitude based (zero-shot) and sensitivity based (single-shot) pruning. We keep the hessian-based criterion for future work.
\end{enumerate}{}

\begin{figure}[t]
\centering
\noindent\begin{tikzpicture}[every node/.style={scale=0.78}]
\def\hspacing{6.3}
\def\vspacing{-4.3}
\node[label=above:{CIFAR10}] (cifar10) at (0,0) {
\begin{tabular}{@{}lrr@{}}
\toprule
\textbf{Model} & \textbf{Param.} & \textbf{Test Err. (\%)} \\ \midrule
CNF-B          & \textgreater{}4M       & 6.44                    \\
CNF-P          & 0.14M                  & 8.03                    \\ \midrule
Maxout~\cite{goodfellow2013maxout} & \textgreater{}5M       & 9.38                        \\
DSN~\cite{lee2015deeply} & 0.97M                  & 7.97                        \\
RCNN-96~\cite{liang2015recurrent} & 0.67M                  & 7.37                        \\
RCNN-160~\cite{liang2015recurrent} & 1.86M                  & 7.09              \\ \bottomrule
\end{tabular}
        };

\node[label=above:{CIFAR100}] (cifar100) at (\hspacing,0) {
\begin{tabular}{@{}lrr@{}}
\toprule
\textbf{Model} & \textbf{Param.} & \textbf{Test Err. (\%)} \\ \midrule
CNF-B          & \textgreater{}4M       & 27.54                   \\
CNF-P          & 0.24M                  & 33.26                   \\ \midrule
Maxout~\cite{goodfellow2013maxout} & \textgreater{}5M       & 38.57                       \\
DSN~\cite{lee2015deeply} & 0.97M                  & 34.57                       \\
RCNN-96~\cite{liang2015recurrent} & 0.67M                  & 34.18                       \\
RCNN-160~\cite{liang2015recurrent} & 1.86M                  & 31.75             \\ \bottomrule
\end{tabular}
        };

\node[label=above:{SVHN}] (svhn) at (0.5*\hspacing,\vspacing) {
\begin{tabular}{@{}lrr@{}}
\toprule
\textbf{Model} & \textbf{Param.} & \textbf{Test Err. (\%)} \\ \midrule
CNF-B          & 0.29M                  & 3.56                    \\
CNF-P          & 15K                    & 4.43                    \\ \midrule
Maxout~\cite{goodfellow2013maxout} & \textgreater{}5M       & 2.47                        \\
DSN~\cite{lee2015deeply} & 0.97M                  & 1.92                        \\
RCNN-160~\cite{liang2015recurrent} & 0.67M                  & 1.80                        \\
RCNN-192~\cite{liang2015recurrent} & 1.86M                  & 1.77          \\ \bottomrule
\end{tabular}
        };

\end{tikzpicture}
\caption{Results on CIFAR10, CIFAR100 and SVHN datasets for various reference methods. CNF-B is our CNF baseline, CNF-P is our best obtained pruned CNF.}
\label{stateofart}
\end{figure}

\section{\label{sec:R}Results on Clean Data}

Due to space limitation, we could not compile the results of all the experiments. An exhaustive listing of the results is provided in the appendix.
%We implemented our experiments in PyTorch, and we trained the models on the Grid5000 platform~\cite{grid5000}. Results are reported in Fig.~\ref{clean-1} and~\ref{clean-2}.

\subsection{State-of-the-Art}

 Fig.~\ref{stateofart} gives an overview of how our CNFs compare to similar state-of-the-art approaches. The pruned CNFs are at par with their competitors where classification error rate is concerned, with significantly smaller networks.
 
 Fig.~\ref{stateofart} does not report results for the PASCALVOC dataset. Since we diverted its data for one-label classification, while it was initially conceived for object detection, it makes our results unfit for making a significant comparison with those reported elsewhere. We would need to compute the Mean Average Precision, based on bounding boxes, for which our CNFs are currently unsuited. However, we believe that it can still be useful to show how our pruned models perform with respect to the baseline CNF on different data distributions.

\subsection{Pruning settings}\label{sec:PS}

Overall, the best models were obtained when applying late pruning, since more training epochs before pruning makes it easier to detect unimportant elements in the network. However, the iterative pruning strategy often gives comparable results to late pruning. We also observe that the magnitude-based (MAGN) criterion gives slightly better results compared to the sensitivity-based (SBD) criterion, as shown in Fig.~\ref{cvsn}. The percentage values are taken from the experiments on all datasets, for all possible settings. For example, in the clean data setting, $18$ networks have been trained on CIFAR$100$: $9$ were pruned using MAGN criterion, and $9$ other using SBD criterion. Thus there are $9$ couples of classifier for which the only difference is the criterion used (all other parameters are the same). Table \ref{cvsn} says that for $83$\% of those couples, the classifier pruned with MAGN was better.

\begin{figure}[t]
\centering
\noindent\begin{tikzpicture}[every node/.style={scale=0.8}]
\def\hspacing{4.5}
\def\vspacing{-4.5}

\node (clean) at (0,0) {
\begin{tabular}{@{}lrr@{}}
\toprule
            & MAGN     & SBD      \\ \midrule
\multicolumn{3}{c}{Clean Setting} \\ \midrule
CIFAR10     & 100\%    & 0\%      \\
CIFAR100    & 83\%     & 17\%     \\
SVHN        & 33\%     & 67\%     \\
PASCALVOC   & 33\%     & 67\%     \\
OVERALL     & 62.5\%     & 37.5\%     \\ \bottomrule
\end{tabular}
        };

\node (noisy) at (\hspacing,0) {
\begin{tabular}{@{}lrr@{}}
\toprule
            & MAGN     & SBD      \\ \midrule
\multicolumn{3}{c}{Noise Setting} \\ \midrule
CIFAR10     & 33\%     & 67\%     \\
CIFAR100    & 50\%     & 50\%     \\
SVHN        & 83\%     & 17\%     \\
PASCALVOC   & 42\%     & 58\%     \\
OVERALL     & 52\%     & 48\%     \\ \bottomrule
\end{tabular}
        };
        
% \node (noisy) at (\hspacing,0) {
% \begin{tabular}{@{}lrr@{}}
% \toprule
%             & MAGN     & SBD      \\ \midrule
% \multicolumn{3}{c}{20\%-Noise Setting} \\ \midrule
% CIFAR10     & 0\%     & 100\%     \\
% CIFAR100    & 67\%     & 33\%     \\
% SVHN        & 100\%     & 0\%     \\
% PASCALVOC   & 0\%     & 100\%     \\
% OVERALL     & 50\%     & 50\%     \\ \bottomrule
% \end{tabular}
%         };

\end{tikzpicture}
\caption{Percentage of the time where the models with the corresponding choices gave the best results.}
\label{cvsn}
\end{figure}

For the training time, the earlier we remove elements of the CNF, the better. Therefore, applying iterative pruning with a MAGN criterion seems to be a good trade-off between loss in accuracy and faster decrease in model complexity, optimizing training time and memory consumption. This is particularly noticeable for pruning sizes of 95\% and 97\%. When pruning 99\% of our models, we saw a clear drop in performance, and the results were too chaotic to base further conclusions on.

In PyTorch, the usual implementation of network pruning does not allow to see the gain in training time, as it is simply done by multiplying the pruned weights with zero. We however have access to a partial gain in training time, thanks to our step of link pruning. Indeed, when we prune the links, we modify directly the architecture of a CNF, thus effectively removing all operations applied by links. This allows to at least show significant improvement in the average training time: 4 hours for early pruning, 6 hours for iterative pruning and 8 hours for late pruning, when the baselines took more than 8 and up to 14 hours. With an implementation of weight pruning that effectively makes use of only the remaining weights, the differences in training time would be far more striking.

\subsection{Link pruning}

\begin{figure}[h]
\centering
\noindent\begin{tikzpicture}
\def\hspacing{6.3}
\def\vspacing{-5}
\def\WIDTH{5cm}
\def\HEIGHT{2.5cm}
\node[label=above:{CIFAR10}] (cifar10) at (0,0) {\includegraphics[width=\WIDTH,height=\HEIGHT]{\WORKDIR/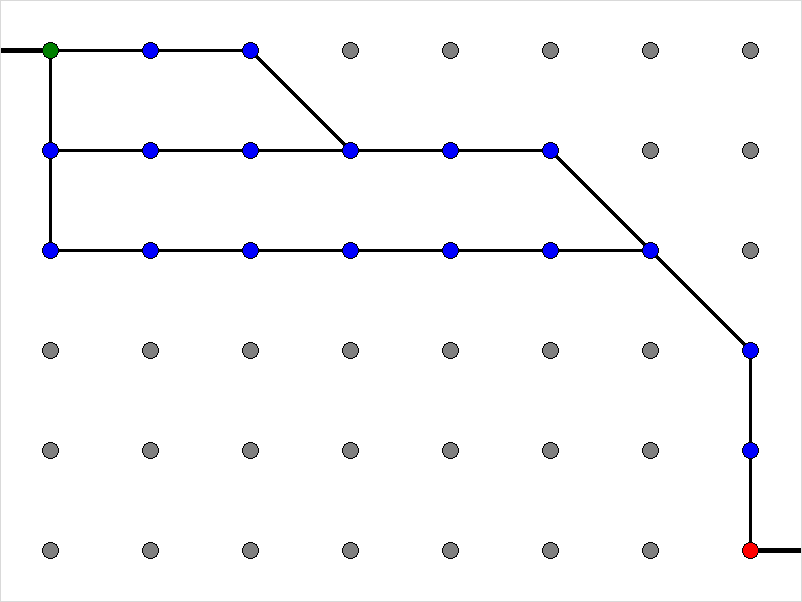}};

\node[label=above:{CIFAR100}] (cifar100) at (\hspacing,0) {\includegraphics[width=\WIDTH,height=\HEIGHT]{\WORKDIR/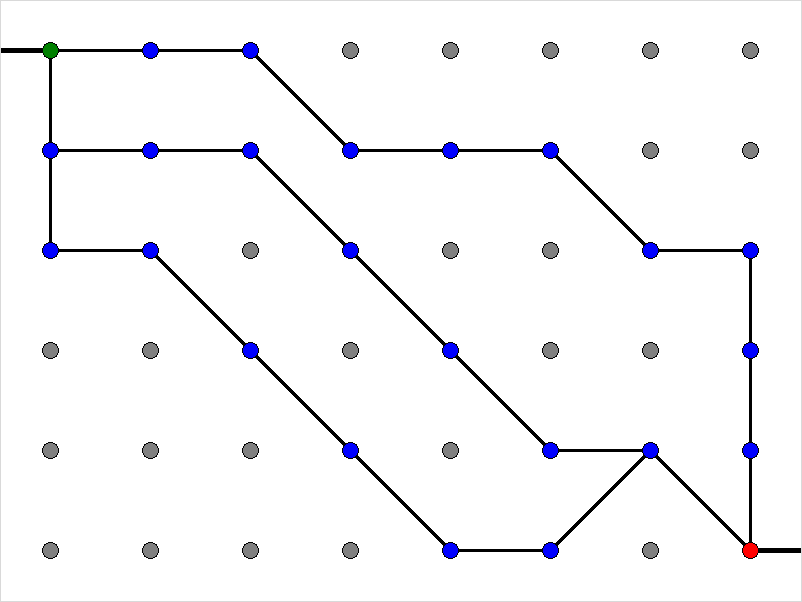}};

% \node[label=above:{SVHN}] (svhn) at (0.5*\hspacing,\vspacing) {\includegraphics[scale=\imscale]{\WORKDIR/IMS/svhn.png}};

\end{tikzpicture}
\caption{Architecture examples of well-performing pruned models: MAGN-ITE-95 for CIFAR10 and CIFAR100%, and MAGN-ITE-97 for SVHN.
}
\label{arch}
\end{figure}

Link pruning allows to have insight in the best performing architectures for each dataset (Fig.~\ref{arch}). We can see for instance that the best for the CIFAR10 and CIFAR100 data seems to be processing it at different scales in parallel in early layers. For CIFAR100, in particular, the best results are obtained when the multiscale processing phase lasts for more layers.

Moreover, we also tried to prune the CNFs without the link pruning step, directly removing weights in the convolution filters. When we compare the models obtained with link pruning with the models obtained without link pruning (table \ref{lp-vs-wp}), we remark that:
\begin{itemize}
    \item when early pruning is applied, adding the link pruning step gives models that clearly outperform those obtained without it;
    \item when iterative or late pruning is applied, the models obtained with or without link pruning have overall similar performance.
\end{itemize}

This indicates that in order to train small models without much resources (computational power, memory, etc), \textit{i.e.} when pruning early on during training is required, applying a link pruning step in the pruning process is clearly beneficial for the performance of the trained models.

\begin{table}[t]
\centering
\caption{Percentage of the time where the models with the corresponding choices gave the best results. {\itshape E.g} for early pruning, link pruning (LP+WP) always outperforms weight pruning (WP): 56\% of the time for MAGN, 44\% of the time for SBD.}
\begin{tabular}{@{}lcc|cc@{}}
\toprule
                   & \multicolumn{2}{c|}{\textbf{MAGN}} & \multicolumn{2}{c}{\textbf{SBD}} \\ \midrule
                   & LP+WP            & WP              & LP+WP           & WP             \\ \midrule
\textbf{EARLY}     & 0.56             & 0.0             & 0.44            & 0.0            \\
\textbf{ITERATIVE} & 0.22              & 0.33            & 0.33             & 0.12           \\
\textbf{LATE}      & 0.45             & 0.45            & 0.10            & 0.0            \\ \bottomrule
\end{tabular}
\label{lp-vs-wp}
\end{table}

\section{Results on Noisy Data\label{sec:noise}}

\subsection{Label Noise}

The last part of this paper consists of evaluating how data annotation quality impacts the outcome of the pruning process. Label noise~\cite{frenay} is known to be harmful to the optimization of supervised classifiers in general~\cite{zhang2003robustness,nettleton2010study} as it biases the training process. But in its broader sense ({\itshape i.e.} an undesired label given to an entity in the data set), is not necessarily limited to training data. It can also be present in validation sets, and even test sets. This has become apparent with the recent works concerning annotation bias~\cite{geva2019modeling}. Therefore, any process making use of data labels is influenced by annotation bias~\cite{lamiroy:hal-01057362}. This could concern network pruning strategies as well, but to the best of our knowledge, no work has been reported on this subject. This work is part of a broader study on the impact of label noise on classifier bias \cite{benje2021}

For our experiments, we will artificially introduce annotation errors in our datasets, so that we can try and compare data-dependent and independent pruning criteria. Furthermore, we want the introduced errors be as close as possible to real world situations, such that the trained models are able to generalize the label noise structure to unseen examples, very much as if the data were mislabelled on subjective interpretations of an annotator, and that this biased interpretation was consequently learned by the machine.

Our goal is to measure if the pruning process exacerbates the bias, or rather remains insensitive to these kinds of noise.

\subsection{Generating Annotation Errors}

Let $\entity$ be an entity in a dataset $\dataset$ belonging to class $\clabel_{\entity}$ with given label $\plabel_{\entity}$. The label predicted for $\entity$ by a given classifier is $\elabel_{\entity}$. We assume that our datasets are void of mislabelled entities:  $\forall \dataset,  \forall \entity \in \dataset, \plabel_{\entity} = \clabel_{\entity}$.

In~\cite{frenay}, annotation errors are defined as the result of a stochastic process $\process$ that disturbs the label of an entity $\entity$ with probability $p_{\entity, \clabel_{\entity}}$. We note $\Tilde{\dataset}$ the result of $\process$ applied to $\dataset$. $\process$ can take three different forms:
\begin{itemize}
	\item uniformly random (Type 1): the probability of an entity being mislabelled does not depend on its class nor on its features ($p_{\entity, \clabel_{\entity}} = p$);
	\item class dependent (Type 2): entities of two different classes can have different probabilities of being mislabelled ($p_{\entity, \clabel_{\entity}} = p_{\clabel_{\entity}}$);
	\item class and feature dependent (Type 3): even for two entities in the same class, the probability that they are mislabelled can vary depending on where they are located in the feature space.
\end{itemize}

Type 1 corresponds to swapping the label of each entity with a fixed probability, and by randomly choosing the noisy label among the remaining ones. For Type 2, one defines a transition matrix between the classes of the problem, such that cell $(i,j)$ represents the probability of labelling an entity of class $i$ with $j$. When this matrix is symmetric, the process is called symmetry flipping, otherwise it is called pair flipping.

As pointed out before, our goal is to find for each dataset $\dataset$, a noisy version $\Tilde{\dataset}$ such that the models trained on $\Tilde{\dataset}$ are effectively biased by the mislabelled entities. To that end, we first obtain a candidate $\Tilde{\dataset} = \process(\dataset)$, we train a baseline CNF on it and we look at two indicators:
\begin{itemize}
	\item \textit{clean fitting}: among the \textit{clean} data samples, the fraction of correctly predicted entities, \textit{i.e.} for which we predicted the \textit{clean} label: $\frac{|\elabel_{\entity} = \clabel_{\entity} = \plabel_{\entity}|}{|\clabel_{\entity} = \plabel_{\entity}|}$
%	\begin{equation}
%	\frac{|\elabel_{\entity} = \clabel_{\entity} = \plabel_{\entity}|}{|\clabel_{\entity} = \plabel_{\entity}|}
%	\end{equation}
	
	\item \textit{noisy fitting}: among the \textit{noisy} data samples, the fraction of entities for which we predicted the \textit{noisy} label:
%	\begin{equation}
	$\frac{|\elabel_{\entity} = \plabel_{\entity}|}{|\clabel_{\entity} \neq \plabel_{\entity}|}$
%	\end{equation}
\end{itemize}

Those indicators are measured for each classifier trained on the noisy data sets, using the test samples (which can also be mislabelled). The clean fitting represents how much a network managed to learn the underlying true task (specified by the remaining correctly labeled entities) without being too much impacted by the mislabelled entities, and the noisy fitting shows how much the network did overfit the label noise structure, and is thus biased by label noise. We consider that a noisy version of a dataset $\dataset$ is acceptable if the networks trained on it obtain high enough values for both indicators:
\begin{itemize}
    \item if the clean fitting is too low, it means the label noise introduced has destroyed too much of the data structure for the classifiers to learn anything, and we do not want to study that setting;
    \item if the noisy fitting is too low, it means the classifiers do not overfit the noisy samples, so the label noise introduced does not have a coherent and learnable structure, while we do believe that real-world label noise is learnable.
\end{itemize}
. 

We empirically determined that uniformly random (Type 1) or class dependant (Type 2) noise types are not satisfactory. Indeed, the noisy fitting indicator remains very low for all networks trained on datasets with such noise types. For our experiment, we therefore decided to only use class and feature dependant noise (Type 3). We generate Type 3 noise for a dataset $\Tilde{\dataset}$ by training an \textit{artificial annotator} $\mathcal{A}$ on the data we want to perturbate. Doing so, we guarantee that the noisy labels effectively depend on the data features. To obtain a given fraction $\epsilon$ of mislabelled samples in $\Tilde{\dataset}$, we stop the training of $\mathcal{A}$ when its test error equals $\epsilon$. In what follows we have generated noisy datasets for CIFAR10, CIFAR100, SVHN and PASCALVOC with $\epsilon=10\%$ and $\epsilon=20\%$.

Networks trained on those noisy datasets obtained higher values for the noisy fitting indicator, confirming the utility of Type 3 noise. Table \ref{cfnf} shows the overall values of the clean and noisy fittings for various networks trained on the four noisy datasets.

\begin{table}[t]
\centering
\caption{Clean and noisy fitting value ranges for models (pruned or not) trained on the noisy versions of each dataset.}
\begin{tabular}{@{}lcccc@{}}
\toprule
              & CIFAR10 & CIFAR100 & SVHN & PASCALVOC \\ \midrule
Clean Fitting & \textgreater{} 80\%    & \textgreater{} 60\%     & $\approx$ 97\% & $\approx$ 50\%      \\
Noisy Fitting & $\approx$ 35\%    & 15\%-25\%     & 40\%-50\% & 10\%-30\%      \\ \bottomrule
\end{tabular}
\label{cfnf}
\end{table}

% We introduce errors in the validation set and test set as well in order to make a comparison also with respect to performance indicators computed on bad quality data.

\subsection{Results}

%The noisy versions of our datasets were obtained by training an \textit{artificial annotator} model on each dataset, and taking its predictions as the new labels for the corresponding data items. More precisely, we trained CNFs and saved their predictions of each model when the test error was equal to 10\%. This way, we built noisy versions of CIFAR10, CIFAR100, SVHN and PASCALVOC datasets with 10\% of mislabelled entities. We also built 20\%-noisy versions of the datasets.

We have repeated the previous experiment conducted on clean data, using this time the noisy datasets. For each data set, we trained $1$ baseline not pruned, and $18$ networks pruned according to the same parameters than before. All networks are biased by label noise in various ways, as the noisy fitting indicator shows in table \ref{cfnf}. The higher the value for one network, the more it did overfit the noisy samples.

First of all, we observed that the pruned networks were in general more biased by label noise than their corresponding baseline, while we were expecting a smaller model to be less prone to overfit the noise. Especially, the models obtained with the MAGN criterion seem to be slightly more biased in general.

Surprisingly, the results were overall better than expected for the SBD criterion, which is data-dependent, and worse than expected for the MAGN criterion, which is data-independent. We were expecting the gap in efficiency between the two criteria to expand when moving from the clean setting to the noisy setting, since we thought that the SBD criterion would prove even worse in presence of label noise, but overall we observed the opposite trend (Fig.~\ref{cvsn}). This may be due to the fact that the more a model overfits the structure of label noise in a dataset, the more it shows through its weights, and therefore through the MAGN criterion as well. The exception is for the SVHN dataset, where the SBD criterion become clearly worse in the noisy setting. However, the dimensions we chose for the models trained on SVHN are far smaller than for the models trained on the other datasets, making them potentially less prone to overfit the label noise structure. This could explain the greater efficiency of the MAGN criterion on SVHN models.

\section{Conclusion}

We showed how to adapt a pruning algorithm to a multidimensional neural network (CNF), obtained results comparable to the state-of-the-art with model sizes $5$ to $50$ times smaller,  and tested the pruning algorithm for many different hyperparameter values. In particular, our pruned networks gave similar performances to those presented in initial CNF paper~\cite{vebreek}, even when we applied an iterative or early-pruning strategy, which allowed to greatly decrease the training time. We showed how including a link pruning step in the pruning process was beneficial to the pruned model, compared to classically applying pruning on the weights directly. We also refuted the initial idea that the presence of label noise in a dataset would make a data-independent criterion better than a data-dependent one, by experimenting on datasets containing $10$\% and $20$\% of mislabelled entities.

\bibliography{egbib}
% \bibliography{sn-bibliography}% common bib file


\begin{appendix}

\section{Appendices}

\begin{figure}[hbtp]
\centering
\includegraphics[scale=0.50]{\WORKDIR/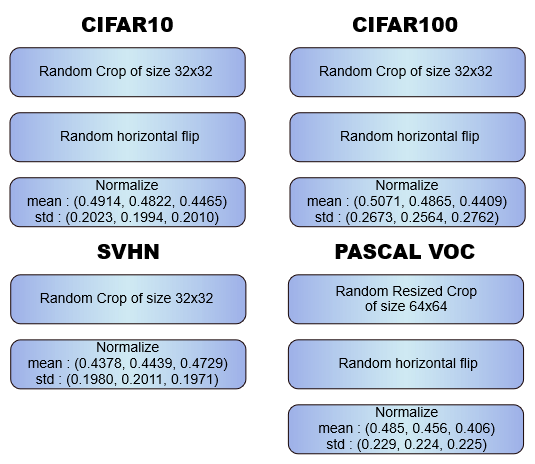}
\caption{Transformations applied to the $4$ training sets for data augmentation.}
\label{augment}
\end{figure}

Following are the results obtained on the $4$ datasets for baseline CNFs (not pruned), and CNFs pruned in various ways, on clean and noisy data. The names of the models represent the different pruning settings: MAGN or SBD for magnitude-based or sensitivity-based pruning; EAR, ITE or LAT for early, iterative or late pruning; and the number represent the fraction of the CNF that has been pruned. For the noisy data setting, the results are still measured on the clean version of the testsets. Also, for all datasets but pascalvoc, and only for the clean data setting, we tried to apply pruning by removing the link pruning step, directly pruning the weights of the models.

\begin{figure}[htbp]
\centering
\noindent\begin{tikzpicture}[every node/.style={scale=0.8}]
\def\hspacing{6.3}
\def\vspacing{-9.5}
\node[label=above:{CIFAR10-CLEAN}] (cifar10) at (0,0) {
\begin{tabular}{@{}lcc@{}}
\toprule
\textbf{Model} & \textbf{Nb. Param.} & \textbf{Test Err. with \textbar{} without LP (\%)} \\ \midrule
BASELINE    & 4523402    & 6.44                           \\ \midrule
MAGN-EAR-95 & 228611     & 09.55 \textbar{} 10.93                    \\
SBD-EAR-95  & -          & 09.94 \textbar{} 10.53                    \\
MAGN-ITE-95 & -          & 08.03 \textbar{} 07.35                    \\
SBD-ITE-95  & -          & 08.99 \textbar{} 07.67                    \\
MAGN-LAT-95 & -          & 08.56 \textbar{} 07.32                    \\
SBD-LAT-95  & -          & 08.69 \textbar{} 08.57                    \\ \midrule
MAGN-EAR-97 & 138194     & 10.49 \textbar{} 12.78                    \\
SBD-EAR-97  & -          & 11.75 \textbar{} 11.31                    \\
MAGN-ITE-97 & -          & 09.97 \textbar{} 08.79                    \\
SBD-ITE-97  & -          & 10.31 \textbar{} 06.39                    \\
MAGN-LAT-97 & -          & 08.39 \textbar{} 08.25                    \\
SBD-LAT-97  & -          & 09.31 \textbar{} 10.40                    \\ \midrule
MAGN-EAR-99 & 47778      & 15.33 \textbar{} 26.56                    \\
SBD-EAR-99  & -          & 14.32 \textbar{} 20.33                    \\
MAGN-ITE-99 & -          & 13.40 \textbar{} 17.11                    \\
SBD-ITE-99  & -          & 16.20 \textbar{} 19.84                    \\
MAGN-LAT-99 & -          & 12.02 \textbar{} 17.83                    \\
SBD-LAT-99  & -          & 14.48 \textbar{} 21.35                    \\ \bottomrule
\end{tabular}
        };

\node[label=above:{CIFAR100-CLEAN}] (cifar100) at (0,\vspacing) {
\begin{tabular}{@{}lcc@{}}
\toprule
\textbf{Model} & \textbf{Nb. Param.} & \textbf{Test Err. with/without LP (\%)} \\ \midrule
BASELINE       & 4529252             & 27.54                                   \\ \midrule
MAGN-EAR-95    & 234461              & 36.41 \textbar{} 38.82                             \\
SBD-EAR-95     & -                   & 37.08 \textbar{} 37.27                             \\
MAGN-ITE-95    & -                   & 34.72 \textbar{} 31.62                             \\
SBD-ITE-95     & -                   & 35.21 \textbar{} 31.77                             \\
MAGN-LAT-95    & -                   & 33.26 \textbar{} 29.95                             \\
SBD-LAT-95     & -                   & 33.73 \textbar{} 33.07                             \\ \midrule
MAGN-EAR-97    & 144044              & 40.98 \textbar{} 42.52                             \\
SBD-EAR-97     & -                   & 39.03 \textbar{} 41.56                             \\
MAGN-ITE-97    & -                   & 34.76 \textbar{} 33.78                             \\
SBD-ITE-97     & -                   & 37.75 \textbar{} 35.77                             \\
MAGN-LAT-97    & -                   & 34.81 \textbar{} 34.09                             \\
SBD-LAT-97     & -                   & 36.39 \textbar{} 37.00                             \\ \midrule
MAGN-EAR-99    & 53628               & 48.71 \textbar{} 57.56                             \\
SBD-EAR-99     & -                   & 47.96 \textbar{} 54.64                             \\
MAGN-ITE-99    & -                   & 42.99 \textbar{} 50.66                             \\
SBD-ITE-99     & -                   & 46.55 \textbar{} 54.13                             \\
MAGN-LAT-99    & -                   & 42.03 \textbar{} 49.09                             \\
SBD-LAT-99     & -                   & 45.92 \textbar{} 53.01                             \\ \bottomrule
\end{tabular}
        };

\end{tikzpicture}
\label{clean-1}
\end{figure}

\newpage

\begin{figure}[h]
\centering
\noindent\begin{tikzpicture}[every node/.style={scale=0.8}]
\def\hspacing{6.3}
\def\vspacing{-9.5}
\node[label=above:{SVHN-CLEAN}] (svhn) at (0,0) {
\begin{tabular}{@{}lcc@{}}
\toprule
\textbf{Model} & \textbf{Nb. Param.} & \textbf{Test Err. with/without LP (\%)} \\ \midrule
BASELINE       & 287594              & 03.56                                   \\ \midrule
MAGN-EAR-95    & 14997               & 05.01/07.90                             \\
SBD-EAR-95     & -                   & 05.20/06.40                             \\
MAGN-ITE-95    & -                   & 04.77/04.77                             \\
SBD-ITE-95     & -                   & 04.43/05.33                             \\
MAGN-LAT-95    & -                   & 05.00/04.82                             \\
SBD-LAT-95     & -                   & 04.47/06.03                             \\ \midrule
MAGN-EAR-97    & 9258                & 06.71/18.94                             \\
SBD-EAR-97     & -                   & 05.74/14.25                             \\
MAGN-ITE-97    & -                   & 06.49/12.73                             \\
SBD-ITE-97     & -                   & 06.46/13.07                             \\
MAGN-LAT-97    & -                   & 04.90/11.46                             \\
SBD-LAT-97     & -                   & 05.90/20.14                             \\ \midrule
MAGN-EAR-99    & 3519                & 14.45/80.41                             \\
SBD-EAR-99     & -                   & 17.77/80.41                             \\
MAGN-ITE-99    & -                   & 80.41/80.41                             \\
SBD-ITE-99     & -                   & 14.54/80.41                             \\
MAGN-LAT-99    & -                   & 17.94/80.41                             \\
SBD-LAT-99     & -                   & 19.88/80.41                             \\ \bottomrule
\end{tabular}
        };
        
\node[label=above:{PASCALVOC-CLEAN}] (pascalvoc) at (0,\vspacing) {
\begin{tabular}{@{}lrr@{}}
\toprule
\textbf{Model} & \textbf{Nb. Param.} & \textbf{Test Err. (\%)}   \\ \midrule
BASELINE             & 5376340          & \textbf{48.78} \\ \midrule
MAGN-EAR-95          & 271876           & 53.44          \\
SBD-EAR-95           & -                & 52.33          \\
MAGN-ITE-95          & -                & \textbf{51.22}          \\
SBD-ITE-95           & -                & 51.69          \\
MAGN-LAT-95          & -                & 51.69          \\
SBD-LAT-95           & -                & 51.64          \\ \midrule
MAGN-EAR-97          & 164413           & 52.70          \\
SBD-EAR-97           & -                & 56.41          \\
MAGN-ITE-97          & -                & 55.46          \\
SBD-ITE-97           & -                & 50.85          \\
MAGN-LAT-97          & -                & 53.07          \\
SBD-LAT-97           & -                & 52.70          \\ \midrule
MAGN-EAR-99          & 56951            & 59.27          \\
SBD-EAR-99           & -                & 84.16          \\
MAGN-ITE-99          & -                & 53.92          \\
SBD-ITE-99           & -                & 54.71          \\
MAGN-LAT-99          & -                & 54.24          \\
SBD-LAT-99           & -                & 55.14          \\ \bottomrule
\end{tabular}
        };
\end{tikzpicture}
\label{clean-2}
\end{figure}

\begin{figure}[t]
\centering
\noindent\begin{tikzpicture}[every node/.style={scale=0.8}]
\def\hspacing{6.3}
\def\vspacing{-9.5}
\node[label=above:{CIFAR10-NOISY-10\%}] (cifar10) at (0,0) {
\begin{tabular}{@{}lrr@{}}
\toprule
\textbf{Model} & \textbf{Nb. Param.} & \textbf{Test Err. (\%)} \\ \midrule
BASELINE       & 4523402             & 13.1                    \\ \midrule
MAGN-EAR-95    & 228611              & 14.37                   \\
SBD-EAR-95     & -                   & 13.81                   \\
MAGN-ITE-95    & -                   & 13.49                   \\
SBD-ITE-95     & -                   & 14.24                   \\
MAGN-LAT-95    & -                   & 13.58                   \\
SBD-LAT-95     & -                   & 13.77                  \\ \midrule
MAGN-EAR-97    & 138194              & 14.53                   \\
SBD-EAR-97     & -                   & 14.59                   \\
MAGN-ITE-97    & -                   & 14.05                   \\
SBD-ITE-97     & -                   & 13.66                   \\
MAGN-LAT-97    & -                   & 13.45                   \\
SBD-LAT-97     & -                   & 13.75                   \\ \midrule
MAGN-EAR-99    & 47778               & 17.35                   \\
SBD-EAR-99     & -                   & 16.99                   \\
MAGN-ITE-99    & -                   & 16.02                   \\
SBD-ITE-99     & -                   & 17.96                   \\
MAGN-LAT-99    & -                   & 15.42                   \\
SBD-LAT-99     & -                   & 17                      \\ \bottomrule
\end{tabular}
        };

\node[label=above:{CIFAR100-NOISY-10\%}] (cifar100) at (0,\vspacing) {
\begin{tabular}{@{}lrr@{}}
\toprule
\textbf{Model} & \textbf{Nb. Param.} & \textbf{Test Err. (\%)} \\ \midrule
BASELINE       & 4529252             & 30.68                   \\ \midrule
MAGN-EAR-95    & 234461              & 39.24                   \\
SBD-EAR-95     & -                   & 39                      \\
MAGN-ITE-95    & -                   & 36.20                   \\
SBD-ITE-95     & -                   & 35.96                   \\
MAGN-LAT-95    & -                   & 35.14                   \\
SBD-LAT-95     & -                   & 35.27                   \\ \midrule
MAGN-EAR-97    & 144044              & 41.33                   \\
SBD-EAR-97     & -                   & 40.84                   \\
MAGN-ITE-97    & -                   & 37.79                   \\
SBD-ITE-97     & -                   & 37.56                   \\
MAGN-LAT-97    & -                   & 36.37                   \\
SBD-LAT-97     & -                   & 37.81                   \\ \midrule
MAGN-EAR-99    & 53628               & 48.17                   \\
SBD-EAR-99     & -                   & 47.98                   \\
MAGN-ITE-99    & -                   & 41.66                   \\
SBD-ITE-99     & -                   & 44.72                   \\
MAGN-LAT-99    & -                   & 43.68                   \\
SBD-LAT-99     & -                   & 47.43                   \\ \bottomrule
\end{tabular}
        };

\end{tikzpicture}
\label{noisy-1}
\end{figure}

\newpage

\begin{figure}[t]
\centering
\noindent\begin{tikzpicture}[every node/.style={scale=0.8}]
\def\hspacing{6.3}
\def\vspacing{-9.5}
\node[label=above:{SVHN-NOISY-10\%}] (svhn) at (0,0) {
\begin{tabular}{@{}lrr@{}}
\toprule
\textbf{Model} & \textbf{Nb. Param.} & \textbf{Test Err. (\%)} \\ \midrule
BASELINE       & 287594              & 7.91                    \\ \midrule
MAGN-EAR-95    & 14997               & 7.79                    \\
SBD-EAR-95     & -                   & 8.1                     \\
MAGN-ITE-95    & -                   & 7.81                    \\
SBD-ITE-95     & -                   & 7.65                    \\
MAGN-LAT-95    & -                   & 7.58                    \\
SBD-LAT-95     & -                   & 7.59                    \\ \midrule
MAGN-EAR-97    & 9258                & 8.59                    \\
SBD-EAR-97     & -                   & 8.41                    \\
MAGN-ITE-97    & -                   & 8.50                    \\
SBD-ITE-97     & -                   & 9.34                    \\
MAGN-LAT-97    & -                   & 8.1                     \\
SBD-LAT-97     & -                   & 8.47                    \\ \midrule
MAGN-EAR-99    & 3519                & 16.22                   \\
SBD-EAR-99     & -                   & 15.9                    \\
MAGN-ITE-99    & -                   & 52.38                   \\
SBD-ITE-99     & -                   & 80.41                   \\
MAGN-LAT-99    & -                   & 14.13                   \\
SBD-LAT-99     & -                   & 19.53                   \\ \bottomrule
\end{tabular}
        };
        
\node[label=above:{PASCALVOC-NOISY-10\%}] (pascalvoc) at (0,\vspacing) {
\begin{tabular}{@{}lrr@{}}
\toprule
\textbf{Model} & \textbf{Nb. Param.} & \textbf{Test Err. (\%)} \\ \midrule
BASELINE       & 5376340             & 52.07                  \\ \midrule
MAGN-EAR-95    & 271876              & 52.86                  \\
SBD-EAR-95     & -                   & 54.13                  \\
MAGN-ITE-95    & -                   & 50.79                  \\
SBD-ITE-95     & -                   & 50.26                  \\
MAGN-LAT-95    & -                   & 52.44                  \\
SBD-LAT-95     & -                   & 51.75                  \\ \midrule
MAGN-EAR-97    & 164413              & 54.13                  \\
SBD-EAR-97     & -                   & 54.08                  \\
MAGN-ITE-97    & -                   & 51.11                  \\
SBD-ITE-97     & -                   & 52.44                  \\
MAGN-LAT-97    & -                   & 52.28                  \\
SBD-LAT-97     & -                   & 52.49                  \\ \midrule
MAGN-EAR-99    & 56951               & 80.83                  \\
SBD-EAR-99     & -                   & 56.73                  \\
MAGN-ITE-99    & -                   & 53.44                  \\
SBD-ITE-99     & -                   & 55.35                  \\
MAGN-LAT-99    & -                   & 53.65                  \\
SBD-LAT-99     & -                   & 53.07                  \\ \bottomrule
\end{tabular}
        };
\end{tikzpicture}
\label{noisy-2}
\end{figure}

\begin{figure}[t]
\centering
\noindent\begin{tikzpicture}[every node/.style={scale=0.8}]
\def\hspacing{6.3}
\def\vspacing{-9.5}
\node[label=above:{CIFAR10-NOISY-20\%}] (cifar10) at (0,0) {
\begin{tabular}{@{}lrr@{}}
\toprule
\textbf{Model} & \textbf{Nb. Param.} & \textbf{Test Err. (\%)} \\ \midrule
BASELINE       & 4523402             & 20.16                   \\ \midrule
MAGN-EAR-95    & 228611              & 19.42                   \\
SBD-EAR-95     & -                   & 18.84                   \\
MAGN-ITE-95    & -                   & 19.58                   \\
SBD-ITE-95     & -                   & 18.16                   \\
MAGN-LAT-95    & -                   & 19.52                   \\
SBD-LAT-95     & -                   & 19.02                  \\ \midrule
MAGN-EAR-97    & 138194              & 18.14                   \\
SBD-EAR-97     & -                   & 18.00                   \\
MAGN-ITE-97    & -                   & 19.11                   \\
SBD-ITE-97     & -                   & 18.95                   \\
MAGN-LAT-97    & -                   & 18.58                   \\
SBD-LAT-97     & -                   & 18.24                   \\ \bottomrule
\end{tabular}
        };

\node[label=above:{CIFAR100-NOISY-20\%}] (cifar100) at (0,\vspacing) {
\begin{tabular}{@{}lrr@{}}
\toprule
\textbf{Model} & \textbf{Nb. Param.} & \textbf{Test Err. (\%)} \\ \midrule
BASELINE       & 4529252             & 38.51                   \\ \midrule
MAGN-EAR-95    & 234461              & 42.66                   \\
SBD-EAR-95     & -                   & 40.98                   \\
MAGN-ITE-95    & -                   & 39.94                   \\
SBD-ITE-95     & -                   & 40.07                   \\
MAGN-LAT-95    & -                   & 38.78                   \\
SBD-LAT-95     & -                   & 40.75                   \\ \midrule
MAGN-EAR-97    & 144044              & 43.69                   \\
SBD-EAR-97     & -                   & 42.65                   \\
MAGN-ITE-97    & -                   & 39.30                   \\
SBD-ITE-97     & -                   & 39.86                   \\
MAGN-LAT-97    & -                   & 40.03                   \\
SBD-LAT-97     & -                   & 42.13                   \\ \bottomrule
\end{tabular}
        };

\end{tikzpicture}
\label{noisy-3}
\end{figure}

\begin{figure}[t]
\centering
\noindent\begin{tikzpicture}[every node/.style={scale=0.8}]
\def\hspacing{6.3}
\def\vspacing{-9.5}
\node[label=above:{SVHN-NOISY-20\%}] (svhn) at (0,0) {
\begin{tabular}{@{}lrr@{}}
\toprule
\textbf{Model} & \textbf{Nb. Param.} & \textbf{Test Err. (\%)} \\ \midrule
BASELINE       & 4523402             & 13.08                   \\ \midrule
MAGN-EAR-95    & 228611              & 11.70                   \\
SBD-EAR-95     & -                   & 11.83                   \\
MAGN-ITE-95    & -                   & 11.84                   \\
SBD-ITE-95     & -                   & 11.85                   \\
MAGN-LAT-95    & -                   & 11.99                   \\
SBD-LAT-95     & -                   & 12.10                  \\ \midrule
MAGN-EAR-97    & 138194              & 11.97                   \\
SBD-EAR-97     & -                   & 12.23                   \\
MAGN-ITE-97    & -                   & 12.36                   \\
SBD-ITE-97     & -                   & 14.79                   \\
MAGN-LAT-97    & -                   & 12.18                   \\
SBD-LAT-97     & -                   & 12.48                   \\ \bottomrule
\end{tabular}
        };

\node[label=above:{PASCALVOC-NOISY-20\%}] (pascalvoc) at (0,\vspacing) {
\begin{tabular}{@{}lrr@{}}
\toprule
\textbf{Model} & \textbf{Nb. Param.} & \textbf{Test Err. (\%)} \\ \midrule
BASELINE       & 4529252             & 56.51                   \\ \midrule
MAGN-EAR-95    & 234461              & 55.99                   \\
SBD-EAR-95     & -                   & 56.14                   \\
MAGN-ITE-95    & -                   & 53.50                   \\
SBD-ITE-95     & -                   & 56.04                   \\
MAGN-LAT-95    & -                   & 58.32                   \\
SBD-LAT-95     & -                   & 54.34                   \\ \midrule
MAGN-EAR-97    & 144044              & 56.57                   \\
SBD-EAR-97     & -                   & 55.30                   \\
MAGN-ITE-97    & -                   & 58.26                   \\
SBD-ITE-97     & -                   & 56.20                   \\
MAGN-LAT-97    & -                   & 57.10                   \\
SBD-LAT-97     & -                   & 54.87                   \\ \bottomrule
\end{tabular}
        };

\end{tikzpicture}
\label{noisy-4}
\end{figure}

\end{appendix}

\end{document}